\documentclass[preprint,12pt]{elsarticle}

\usepackage{amsmath,amsfonts}
\usepackage{algorithmic}
\usepackage{array}
\usepackage[tight,footnotesize]{subfigure}
\usepackage{textcomp}
\usepackage{stfloats}
\usepackage{url}
\usepackage{verbatim}
\usepackage{graphicx}
\def\BibTeX{{\rm B\kern-.05em{\sc i\kern-.025em b}\kern-.08em
    T\kern-.1667em\lower.7ex\hbox{E}\kern-.125emX}}
\usepackage{balance}
\usepackage{booktabs}
\usepackage{caption}
\usepackage{makecell}
\usepackage{adjustbox}
\usepackage{wrapfig}
\usepackage[linesnumbered,ruled]{algorithm2e}
\usepackage{amssymb}
\usepackage{lineno}
\usepackage{float}
\usepackage{algorithmic}
\usepackage[tight,footnotesize]{subfigure}
\usepackage{booktabs}
\usepackage{caption}
\usepackage{makecell}
\usepackage{adjustbox}
\usepackage{wrapfig}
\usepackage{xcolor}
\usepackage[linesnumbered,ruled]{algorithm2e}
\usepackage{listings}
\usepackage{makecell}
\newcommand{\cmark}{\ding{51}}%
\newcommand{\xmark}{\ding{55}}%

\journal{Journal Name}

\begin{document}
\sloppy
\setlength{\parskip}{0pt}

\begin{frontmatter}




\title{Toward Zero-Egress Psychiatric AI: On-Device LLM Deployment for Privacy-Preserving Mental Health Decision Support}


\author[label1]{Eranga Bandara}
\ead{cmedawer@odu.edu}
\author[label10]{Asanga Gunaratna}
\ead{founders@complianceoslab.app}
\author[label1]{Ross Gore}
\ead{rgore@odu.edu}
\author[label4]{Anita H.\ Clayton}
\ead{AHC8V@uvahealth.org}
\author[label1]{Christopher K.\ Rhea}
\ead{crhea@odu.edu}
\author[label7]{Sachini Rajapakse}
\ead{sachini.rajapakse@iciclelabs.ai}
\author[label7]{Isurunima Kularathna}
\ead{isurunima.kularathna@iciclelabs.ai}
\author[label1]{Sachin Shetty}
\ead{sshetty@odu.edu}
\author[label1]{Ravi Mukkamala}
\ead{mukka@odu.edu}
\author[label5]{Xueping Liang}
\ead{xuliang@fiu.edu}
\author[label3]{Preston Samuel}
\ead{preston.l.samuel.mil@health.mil}
\author[label2]{Atmaram Yarlagadda}
\ead{atmaram.yarlagadda.civ@health.mil}

\address[label1]{Old Dominion University, Norfolk, VA, USA}
\address[label10]{AI Motion Labs, Melbourne, Australia}
\address[label4]{Department of Psychiatry and Neurobehavioral Sciences, \\ University of Virginia School of Medicine, Charlottesville, VA, USA}
\address[label7]{IcicleLabs.AI}
\address[label5]{Florida International University, FL, USA}
\address[label3]{Blanchfield Army Community Hospital, Fort Campbell, KY, USA}
\address[label2]{McDonald Army Health Center, Newport News, VA, USA}

\begin{abstract}
Privacy represents one of the most critical yet underaddressed barriers to
AI adoption in mental healthcare---particularly in high-sensitivity
operational environments such as military, correctional, and remote
healthcare settings, where the risk of patient data exposure can deter
help-seeking behavior entirely.
Existing AI-enabled psychiatric decision support systems predominantly rely
on cloud-based inference pipelines, requiring sensitive patient data to
leave the device and traverse external servers, creating unacceptable
privacy and security risks in these contexts.
In this paper, we propose a zero-egress, on-device AI platform for
privacy-preserving psychiatric decision support, deployed as a
cross-platform mobile application.
The proposed system extends our prior work on fine-tuned LLM consortiums for psychiatric diagnosis
standardization by fundamentally
re-architecting the inference pipeline for fully local execution---ensuring
that no patient data is transmitted to, processed by, or stored on any
external server at any stage.
The platform integrates a consortium of three lightweight, fine-tuned, and
quantized open-source LLMs---Gemma (Google DeepMind),
Phi-3.5-mini (Microsoft), and
Qwen2 (Alibaba)---selected for their compact
architectures and proven efficiency on resource-constrained mobile hardware.
Each model is fine-tuned on curated psychiatrist--patient conversational
datasets and compressed using Quantized Low-Rank
Adaptation~(QLoRA) to enable real-time on-device
inference.
An on-device orchestration layer coordinates ensemble inference and
consensus-based diagnostic reasoning, producing DSM-5-aligned assessments
for conditions including depression, anxiety disorder, post-traumatic stress
disorder~(PTSD), and schizophrenia.
The platform is designed to assist clinicians with differential diagnosis
and evidence-linked symptom mapping, as well as to support patient-facing
self-screening with appropriate clinical safeguards.
Initial evaluation demonstrates that the proposed zero-egress deployment
achieves diagnostic accuracy comparable to its server-side predecessor
while sustaining real-time inference latency on commodity mobile hardware.
To the best of our knowledge, this work represents the first fully
on-device, zero-egress psychiatric AI platform built on a fine-tuned LLM
ensemble, establishing a deployable and privacy-sound foundation for
AI-augmented mental healthcare in operationally sensitive and
resource-constrained environments.
\end{abstract}


\end{frontmatter}

\section{Introduction}

Mental health disorders represent one of the most pressing and
underserved challenges in global public health. According to the World
Health Organization, nearly one billion people worldwide live with a mental
disorder, yet more than two-thirds of those in low- and middle-income
countries receive no treatment whatsoever~\cite{who2022mental}. Even in
high-income countries, significant gaps persist between the prevalence of
psychiatric conditions and access to timely, accurate clinical evaluation.
The shortage of trained psychiatrists, the subjectivity inherent to
diagnosis, cultural and linguistic barriers, and---critically---the pervasive
stigma surrounding mental illness collectively impede care delivery at
scale~\cite{Saxena2007, Thornicroft2016}.

Among the structural barriers to mental healthcare, privacy is
particularly acute in high-sensitivity operational contexts. Military
personnel, veterans, correctional populations, and individuals in remote
or underserved settings face compounded obstacles: not only is access to
psychiatric professionals limited, but the perceived risk of disclosure can
itself deter help-seeking behavior~\cite{Hoge2004, Kim2011}. Studies have
shown that military service members are significantly less likely to seek
mental health treatment when they believe their data may be accessed by
commanding officers or entered into institutional records~\cite{Greene2010}.
Similar dynamics affect patients in rural clinics, humanitarian field
deployments, and other environments where data sovereignty cannot be
assumed. In these contexts, any AI-assisted psychiatric tool that routes
sensitive conversational data through external servers introduces not merely
a technical privacy risk, but a real-world barrier to adoption that undermines
the very access it aims to improve.

Existing AI-enabled decision support systems for mental health have
demonstrated considerable promise in augmenting clinical
judgment~\cite{Guo2020, Shim2021}. Recent work has shown that
LLMs fine-tuned on psychiatrist--patient
conversational data can achieve high accuracy in identifying conditions
such as depression, anxiety, post-traumatic stress disorder~(PTSD), and
schizophrenia, with outputs that align closely with DSM-5 diagnostic
criteria~\cite{bandara2025standardization}. However, virtually all existing
platforms---including our own prior work---depend on cloud-based or
server-side inference pipelines in which patient conversational data is
transmitted to remote compute infrastructure for
processing~\cite{bandara2025standardization, Guo2020}. This architectural
dependency creates an irreconcilable tension with the privacy requirements
of the most at-risk and underserved populations.

The emergence of lightweight, highly efficient open-source LLMs---such
as Gemma~\cite{team2024gemma}, Phi-3.5-mini~\cite{abdin2024phi3}, and
Qwen2~\cite{yang2024qwen2}---combined with advances in model compression
via Quantized Low-Rank Adaptation~(QLoRA)~\cite{dettmers2023qlora} and
mobile inference runtimes, has fundamentally changed what is computationally
achievable on consumer mobile hardware. Modern smartphones equipped with
dedicated neural processing units~(NPUs) are now capable of running
billion-parameter language models at interactive speeds, opening the door
to genuinely on-device, zero-egress AI
inference~\cite{xu2024survey, laskaridis2024}. This technological shift
creates a concrete opportunity to re-architect psychiatric decision support
systems such that no patient data ever leaves the device---not during
inference, not for logging, and not for model updates.

In this paper, we present a zero-egress, on-device AI platform for
privacy-preserving psychiatric decision support, deployed as a cross-platform
mobile application. The proposed system directly extends prior work on
server-side fine-tuned LLM consortiums for psychiatric diagnosis
standardization~\cite{bandara2025standardization}, in which we demonstrated
that a consortium of fine-tuned LLMs---coordinated by a reasoning
model---can produce accurate, DSM-5-aligned psychiatric diagnoses from
natural language clinical conversations. The present work addresses the
fundamental privacy limitation of that architecture by re-engineering the
entire inference pipeline for fully local, on-device execution. Three
lightweight LLMs---Gemma, Phi-3.5-mini, and Qwen2---are fine-tuned on
curated psychiatrist--patient conversational datasets, quantized using QLoRA
for mobile deployment, and orchestrated by an on-device ensemble reasoning
layer that produces consensus-based diagnostic assessments without any data
egress. The platform is integrated within a cross-platform mobile clinical
decision support application and is designed to serve both clinicians seeking
differential diagnosis support and patients engaging in guided self-screening.

The primary contributions of this work are as follows:

\begin{enumerate}

    \item \textbf{Zero-egress psychiatric AI architecture:} We propose and
    implement a fully on-device inference architecture for psychiatric
    decision support in which patient data is never transmitted to or
    processed by any external server, establishing a privacy-by-design
    foundation for clinical AI in sensitive operational environments.

    \item \textbf{Lightweight LLM fine-tuning for mobile deployment:} We
    fine-tune a consortium of three compact, open-source LLMs---Gemma,
    Phi-3.5-mini, and Qwen2---on domain-specific psychiatrist--patient
    conversational datasets and demonstrate their viability for real-time
    psychiatric reasoning on consumer mobile hardware via QLoRA-based
    compression.

    \item \textbf{On-device ensemble reasoning:} We adapt the consensus-based
    multi-model diagnostic reasoning framework from our prior server-side
    work~\cite{bandara2025standardization} for fully local execution,
    replacing the cloud-hosted reasoning LLM with an on-device orchestration
    layer that aggregates and reconciles ensemble predictions without
    external dependencies.

    \item \textbf{Cross-platform mobile clinical application:} We integrate
    the zero-egress inference pipeline within a cross-platform mobile
    application that supports both clinician-facing differential diagnosis
    assistance and patient-facing guided self-screening, with appropriate
    clinical safeguards aligned with DSM-5 standards.


\end{enumerate}

To the best of our knowledge, this work represents the first fully
on-device, zero-egress psychiatric AI platform built on a fine-tuned LLM
ensemble, and the first to address the intersection of mobile edge inference,
clinical diagnostic accuracy, and privacy-preserving design specifically
within the domain of mental health decision support.

The remainder of this paper is organized as follows.
Section 2 provides background on lightweight LLMs,
model compression for mobile deployment, and privacy-preserving AI in
healthcare.
Section 3 describes the overall system architecture
and zero-egress design principles.
Section 4 details the on-device LLM fine-tuning and
deployment pipeline.
Section 5 presents the clinical functionality and
mobile application design.
Section 6 reports implementation details and evaluation
results.
Section 7 reviews related work.
Section 8 concludes the paper and discusses future
directions.


\section{Background}
\label{sec:background}

This section provides the foundational context for the proposed platform,
covering five interconnected areas: the psychiatric diagnosis process and
its inherent limitations; the global mental health access crisis and the
specific barriers faced by privacy-sensitive populations; the emergence of
LLMs in clinical decision support; advances in lightweight
model architectures and compression techniques that enable on-device
deployment; and the growing body of work on privacy-preserving AI in
healthcare.

\subsection{Psychiatric Diagnosis: Process, Subjectivity, and Variability}

Psychiatric diagnosis is fundamentally a language-driven process. Unlike
most branches of medicine, psychiatry lacks definitive biomarkers, imaging
signatures, or laboratory tests that can objectively confirm the presence of
a disorder. Instead, clinicians rely on structured clinical interviews,
behavioral observation, and patient self-report, interpreted through
standardized frameworks such as the Diagnostic and Statistical Manual of
Mental Disorders, Fifth Edition~(DSM-5)~\cite{APA2013} and the
International Classification of Diseases~(ICD-11)~\cite{WHO2019ICD}.

This reliance on naturalistic dialogue introduces substantial variability.
Inter-rater reliability --- the degree to which two independent clinicians
reach the same diagnosis for the same patient --- has been a persistent
concern in psychiatric research. Studies have reported kappa coefficients
as low as 0.2--0.4 for several diagnostic categories, indicating only fair
to moderate agreement even among trained
specialists~\cite{Regier2013, Freedman2013}. Variability arises from
multiple sources: differences in clinical training and experience,
cultural and linguistic factors that shape symptom expression and
interpretation, time constraints in clinical encounters, and the
considerable overlap in symptom profiles across diagnostic
categories~\cite{Kendler2009}. Conditions such as depression, bipolar
disorder, generalized anxiety disorder, and PTSD share many surface-level
features---fatigue, sleep disruption, concentration difficulties---and
distinguishing between them requires careful, systematic inquiry that is
difficult to guarantee under real-world clinical conditions.

The consequences of diagnostic variability are clinically significant.
Misdiagnosis or delayed diagnosis can result in inappropriate treatment,
prolonged suffering, and in severe cases, preventable harm. For conditions
with episodic presentations such as bipolar disorder, delays in accurate
diagnosis have been documented at an average of six to ten
years~\cite{Hirschfeld2003}. These gaps underscore the need for
decision support tools that can bring consistency, structure, and
evidence-based reasoning to the diagnostic process without supplanting
clinical judgment.

\subsection{Mental Health Access and Privacy-Sensitive Populations}

The global mental health treatment gap---the proportion of people with a
diagnosable disorder who receive no care---remains unacceptably wide.
The World Health Organization estimates that treatment gaps exceed 70\%
in most low- and middle-income countries, and remain significant even in
high-income settings~\cite{who2022mental}. Multiple intersecting factors
drive this gap, including workforce shortages~\cite{Saxena2007}, geographic
inaccessibility, cost, and the pervasive stigma that surrounds mental
illness in most cultures~\cite{Thornicroft2016}.

Among populations for whom privacy is a structural concern rather than
merely a personal preference, access barriers are compounded in ways that
standard clinical delivery models do not adequately address. Military
personnel represent perhaps the most well-documented example. Surveys
consistently show that service members with diagnosable mental health
conditions --- particularly depression, PTSD, and alcohol use disorder ---
avoid seeking care at rates significantly higher than the general
population~\cite{Hoge2004}. The primary barrier cited is not distance or
cost, but fear of stigma and, critically, concern that disclosure will
affect career progression, security clearances, or unit
cohesion~\cite{Kim2011, Greene2010}. This creates a structural problem:
the populations most exposed to psychological trauma are the least likely
to seek care from systems that record and transmit their disclosures~\cite{deep-psychiatric}.

Similar dynamics play out in correctional settings, where incarcerated
individuals may avoid mental health engagement due to fears about how
disclosures are used in judicial or parole proceedings. In humanitarian
and disaster response contexts, displaced persons and refugees may have
well-founded distrust of institutional data systems entirely. In each
of these settings, the architecture of a mental health AI tool --- and
specifically, whether patient conversational data leaves the device ---
is not a secondary technical consideration, but a primary determinant
of whether the tool will be used at all. This insight motivates the
zero-egress design at the core of the proposed platform.

\subsection{LLMs in Clinical Decision Support}

LLMs are transformer-based deep learning models
trained on large corpora of text to perform a wide range of natural language
understanding and generation tasks~\cite{Vaswani2017, Brown2020}. Their
ability to interpret conversational language, reason over multi-turn
dialogues, and generate structured outputs has made them compelling
candidates for clinical applications in which natural language is the
primary medium of interaction~\cite{proof-of-tbi}.

In the domain of mental health, LLMs have been applied to depression
screening~\cite{Guo2020}, suicide risk
assessment~\cite{Gaur2021}, psychotherapy dialogue
analysis~\cite{Flemotomos2021}, and standardized symptom
assessment~\cite{Shim2021}. A key finding across this body of work is
that general-purpose LLMs, while capable of identifying broad symptom
patterns, tend to produce unstructured and clinically imprecise outputs
when applied to diagnostic tasks. Fine-tuning on domain-specific
clinical dialogue datasets substantially improves both the accuracy and
the clinical structure of model outputs, enabling DSM-aligned diagnostic
reasoning from free-text conversational
input~\cite{bandara2025standardization}.

Our prior work demonstrated this effect directly: a consortium of
fine-tuned LLMs --- Llama-3, Mistral, and Qwen2 --- trained on
approximately 2,000 annotated psychiatrist--patient conversations achieved
significantly improved diagnostic precision across multiple conditions,
with a consensus-driven reasoning layer further refining ensemble
outputs~\cite{bandara2025standardization}. However, that architecture
relied on server-side inference and an external reasoning model
(OpenAI-gpt-oss), making it unsuitable for contexts in which data
cannot leave the device. The present work addresses this gap directly.

Beyond accuracy, LLMs in clinical settings raise important questions
about explainability, bias, and responsible deployment. Models fine-tuned
on clinical data may inherit biases present in that data, including
under-representation of certain demographic groups, cultural variation
in symptom expression, and the diagnostic assumptions of the clinicians
who generated the training labels~\cite{Chen2021bias}. Any clinical
deployment of LLM-based decision support must therefore include
mechanisms for clinician oversight, uncertainty communication, and
regular model auditing --- considerations that inform the design of the
proposed platform.

\subsection{Lightweight LLMs and On-Device Deployment}

Until recently, the deployment of LLMs on mobile hardware was
computationally infeasible. Models of meaningful capability typically
required tens of gigabytes of memory and GPU compute far exceeding that
available on smartphones. This landscape has shifted significantly since
2023, driven by two parallel developments: the emergence of purpose-built
lightweight model architectures, and advances in post-training compression
techniques.

On the architecture side, models such as Google DeepMind's
Gemma~\cite{team2024gemma}, Microsoft's Phi series~\cite{abdin2024phi3},
and Alibaba's Qwen2~\cite{yang2024qwen2} have demonstrated that
carefully curated training data and architectural optimizations can yield
models in the 1B--4B parameter range that match or exceed the performance
of much larger models on targeted downstream tasks. Gemma, for instance,
leverages distillation from the Gemini model family and is explicitly
designed for on-device deployment via Google's MediaPipe/LiteRT
runtime~\cite{team2024gemma}. Phi-3.5-mini achieves strong reasoning
performance at 3.8B parameters through training on high-quality,
reasoning-dense data, and is available in ONNX format optimized for
mobile inference~\cite{abdin2024phi3}. Qwen2 offers a range of sizes
from 0.5B to 72B, with the smaller variants explicitly designed for
multilingual, privacy-preserving on-device use~\cite{yang2024qwen2, deep-stride}.

On the compression side, QLoRA has emerged
as the dominant technique for adapting and compressing LLMs for
resource-constrained environments~\cite{dettmers2023qlora}. QLoRA
combines 4-bit quantization of base model weights with the insertion of
trainable low-rank adapter matrices, enabling fine-tuning and inference
at a fraction of the memory cost of full-precision models. At 4-bit
quantization, a 3B parameter model requires approximately 1.5--2~GB of
memory --- well within the capacity of modern flagship and mid-range
smartphones, which typically offer 6--12~GB of RAM. Further compression
via GGUF~\cite{gguf2023} format and frameworks such as
llama.cpp~\cite{llamacpp2023} and MLC LLM~\cite{mlcllm2023} enables
efficient token generation on mobile CPUs and NPUs without requiring
dedicated GPU hardware.

Empirical studies of on-device LLM inference have confirmed the viability
of this approach on real consumer hardware. Laskaridis~\textit{et al.}
demonstrated that sub-4B quantized models can achieve interactive
inference speeds~(5--15 tokens per second) on modern
smartphones~\cite{laskaridis2024}, and Xu~\textit{et al.} provide a
comprehensive survey of resource-efficient architectures suitable for
mobile deployment~\cite{xu2024survey}. These results establish the
technical foundation on which the proposed platform is built.

\subsection{Privacy-Preserving AI in Healthcare}

The application of AI in healthcare is tightly constrained by regulatory
frameworks designed to protect patient data. In the United States, the
Health Insurance Portability and Accountability Act~(HIPAA) mandates
strict controls over the handling of Protected Health Information~(PHI),
including conversational data generated during clinical
encounters~\cite{hipaa1996}. In the European Union, the General Data
Protection Regulation~(GDPR) imposes similarly stringent requirements,
with health data classified as a special category subject to the highest
level of protection~\cite{GDPR2016}. For military and government
contexts, additional frameworks --- including DoD Instruction 8582.01
on privacy in the DoD~\cite{DoDI8582} and FedRAMP cloud security
standards~\cite{FedRAMP} --- further constrain the architectures
permissible for clinical AI deployment.

Cloud-based AI systems, in which patient data is transmitted to and
processed on remote servers, face significant compliance challenges under
these frameworks. While cloud providers offer HIPAA Business Associate
Agreements and FedRAMP-authorized environments, these arrangements
introduce latency, dependency on network availability, and residual
exposure risk from data in transit and at rest on third-party
infrastructure~\cite{Blobel2018}. For psychiatric data specifically,
the sensitivity is heightened: mental health records have historically
been subject to stronger confidentiality protections than general medical
records in many jurisdictions, reflecting the particular harms that can
flow from their unauthorized disclosure~\cite{Appelbaum2015}.

On-device, zero-egress inference eliminates the data transmission
surface entirely. Patient conversational data never leaves the device,
and no network connection is required for inference --- making the system
suitable for offline deployment in field, remote, or operationally
restricted environments. This approach represents the strongest available
form of privacy-by-design for clinical AI: protection is architectural
rather than contractual, and does not depend on the security practices
of any third party. Recent work in federated learning~\cite{Rieke2020}
and local differential privacy~\cite{Duchi2013} has advanced the
theoretical foundations of privacy-preserving AI in healthcare, but
on-device inference remains the most practically deployable form of
zero-egress AI for individual clinical interactions, and is the approach
adopted in this work.

\section{System Architecture}
\label{sec:architecture}

This section presents the architecture of the proposed zero-egress,
on-device psychiatric AI platform. We first articulate the core design
principles that governed all architectural decisions, followed by a
high-level overview of the three-layer stack. We then describe each
layer in detail, and conclude with the privacy and security architecture
that enforces the zero-egress guarantee throughout the system.

\subsection{Design Principles}

The architecture of the proposed platform is governed by three
foundational design principles that collectively enforce privacy,
accessibility, and responsible clinical use.

\textbf{Privacy by architecture, not by policy.} The strongest available
privacy guarantee for sensitive clinical data is one that is structural
rather than contractual. Cloud-based systems depend on the security
practices, legal agreements, and infrastructure configurations of third
parties to protect patient data. The proposed platform eliminates this
dependency entirely: all inference, orchestration, and storage operations
execute on the user's device, and patient conversational data never
traverses a network interface at any point. This zero-egress constraint
is not a configurable option---it is baked into the system's
architecture.

\textbf{Offline-first operation.} The target deployment contexts for
this platform---military field environments, rural clinics, humanitarian
deployments, correctional facilities---frequently involve unreliable or
absent network connectivity. The platform is designed to function with
full capability in the absence of any network connection. Model weights
are stored locally, inference is performed on-device, and clinical
outputs are generated without requiring any external service call. Network
connectivity, when present, is used only for optional, user-initiated
actions such as clinician-authorized record export.

\textbf{Clinician-in-the-loop.} The platform is explicitly designed to
augment clinical judgment rather than replace it. All diagnostic outputs
are presented as decision support --- structured suggestions, confidence
indicators, and DSM-5 criteria mappings --- rather than definitive
diagnoses. Final clinical determination remains with the licensed
practitioner. Patient-facing outputs are further filtered through an
additional safeguard layer that withholds raw diagnostic labels and
instead provides structured self-assessment feedback with escalation
prompts where appropriate.

\subsection{High-Level Architecture Overview}

\begin{figure}[H]
\centering
\includegraphics[width=4.6in]{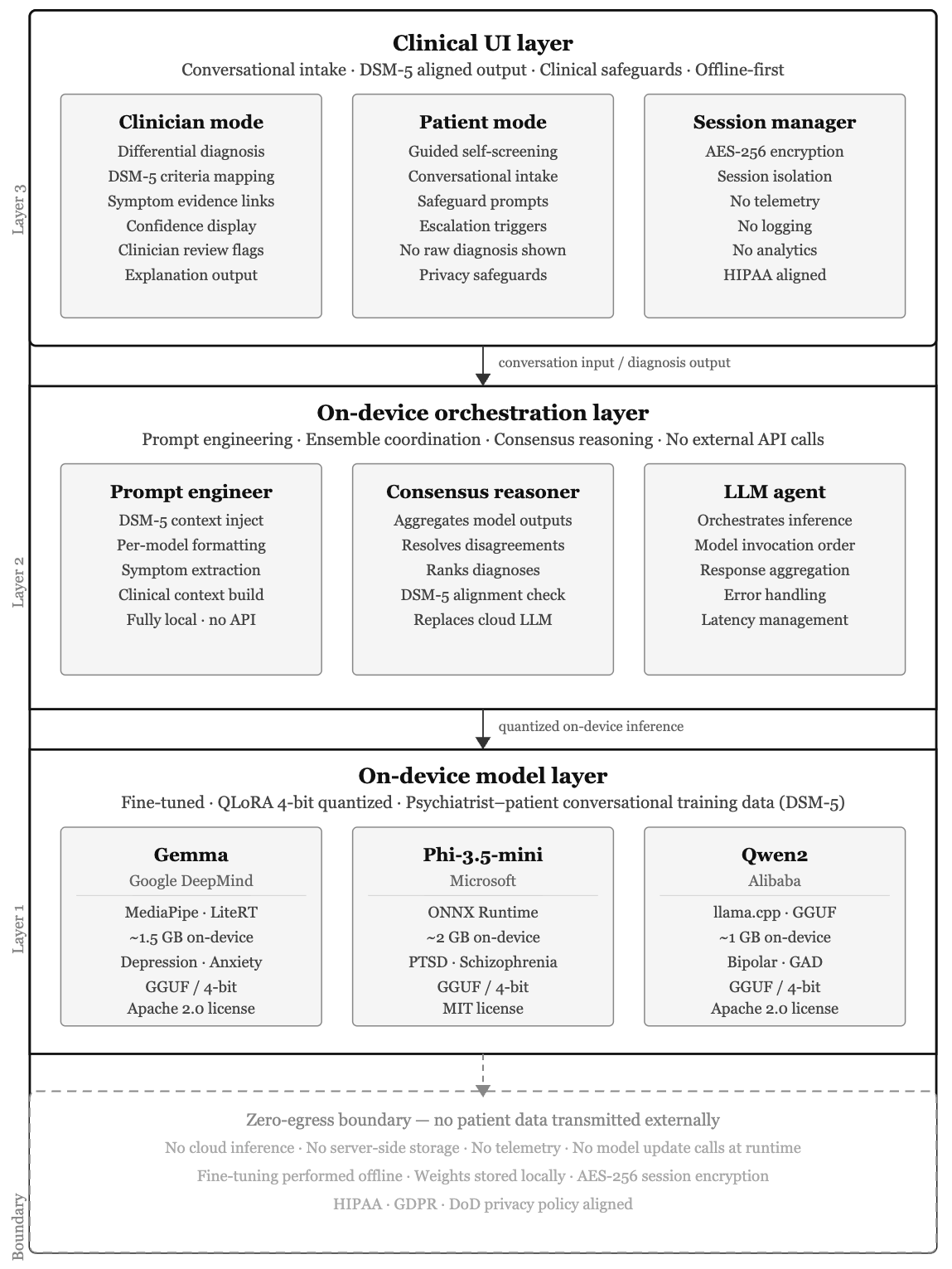}
\DeclareGraphicsExtensions.
\vspace{-0.1in}
\caption{Zero-egress on-device architecture for privacy-preserving
psychiatric AI decision support. Layer~1 houses the on-device model
layer, comprising three fine-tuned, QLoRA 4-bit quantized open-source
LLMs (Gemma, Phi-3.5-mini, and Qwen2), each fine-tuned on
psychiatrist--patient conversational datasets aligned with DSM-5
diagnostic criteria. Layer~2 provides the on-device orchestration layer,
coordinating prompt engineering, ensemble inference, and consensus-based
diagnostic reasoning without any external API dependency. Layer~3
presents the clinical UI layer, offering separate clinician and patient
interaction modes with embedded clinical safeguards and session-level
privacy controls. The dashed boundary at the base denotes the zero-egress
enforcement zone: no patient data is transmitted externally at any stage
of the inference or storage pipeline.}
\label{fig:architecture}
\end{figure}

The platform is organized into three layers, all executing entirely
within the mobile device boundary, plus a zero-egress enforcement
boundary that explicitly prohibits any outbound data transmission.
Figure~\ref{fig:architecture} illustrates the complete architecture.

\subsection{Layer 1 --- On-Device Model Layer}

The on-device model layer forms the analytical core of the platform. It
comprises a consortium of three lightweight, open-source LLMs, each
fine-tuned on domain-specific psychiatrist--patient conversational
datasets and compressed for efficient execution on consumer mobile
hardware.

\textbf{Model selection.} The three models --- Gemma~\cite{team2024gemma},
Phi-3.5-mini~\cite{abdin2024phi3}, and Qwen2~\cite{yang2024qwen2} ---
were selected on the basis of four criteria: parameter efficiency
(sub-4B parameters suitable for mobile deployment), demonstrated
reasoning capability on clinical language tasks, permissive open-source
licensing (Apache 2.0 or MIT), and availability of mobile-optimized
inference runtimes~\cite{astride}. Together, the three models contribute diverse
architectural inductive biases to the ensemble, improving the robustness
of consensus-based diagnostic reasoning.

\textbf{Quantization and compression.} Each model is compressed using
QLoRA~\cite{dettmers2023qlora} with
4-bit weight quantization, reducing the on-device memory footprint to
approximately 1.0--2.0~GB per model. Gemma is deployed via Google's
MediaPipe~/ LiteRT runtime, which is optimized for mobile NPU
acceleration. Phi-3.5-mini is packaged in ONNX format and executed via
ONNX Runtime Mobile, enabling cross-platform deployment on both iOS and
Android. Qwen2 is quantized to GGUF format and executed via
llama.cpp~\cite{llamacpp2023}, a highly portable inference engine with
broad mobile hardware support.

\textbf{Fine-tuning.} All three models were fine-tuned offline --- prior
to deployment --- on a curated dataset of approximately 2,000 annotated
psychiatrist--patient conversational records, each containing a clinical
conversation, diagnostic reasoning trace, and a DSM-5 aligned final
diagnosis. Fine-tuning was performed using the Unsloth
library~\cite{unsloth2024} on GPU infrastructure and is
not performed at runtime. The resulting fine-tuned, quantized weights are
packaged with the application and stored in local encrypted storage on
the device.

\subsection{Layer 2 --- On-Device Orchestration Layer}

The orchestration layer coordinates the full inference pipeline without
any external API calls or network dependencies. It comprises three
components: a prompt engineer, a consensus reasoner, and an LLM agent.

\textbf{Prompt engineer.} The prompt engineer constructs model-specific
input prompts from the raw clinical conversation captured by the UI
layer. For each model in the ensemble, a tailored prompt is generated
that embeds the conversational input alongside structured DSM-5
diagnostic context, including relevant symptom categories, duration
criteria, and functional impairment indicators. Prompt templates are
designed to elicit structured diagnostic outputs --- including a
candidate diagnosis, associated DSM-5 code, confidence indicator, and
supporting symptom evidence --- from each fine-tuned model~\cite{agentic-ai, agentic-workflow-practicle-guide}.

\textbf{LLM agent.} The LLM agent manages the invocation of each model
in the ensemble, handling inference scheduling, response collection, and
error management. The agent executes model calls sequentially or in
parallel depending on available device resources, and structures the
individual model outputs into a unified format for downstream processing
by the consensus reasoner~\cite{agent-survey, agentsway}.

\textbf{Consensus reasoner.} The consensus reasoner implements a fully
local reasoning mechanism that coordinates the diagnostic outputs of the
three fine-tuned models without any external API dependency.
It receives the structured diagnostic outputs from
all three models and applies a weighted consensus algorithm
to identify the most consistent and clinically appropriate diagnosis~\cite{towards-rai-xai}.
Disagreements between models are resolved through a combination of
majority voting, confidence-weighted aggregation, and DSM-5 criterion
alignment checks. The final output --- a ranked list of candidate
diagnoses with supporting evidence and DSM-5 code mappings --- is passed
to the UI layer for presentation~\cite{agentic-ai-transition-organization}.

\subsection{Layer 3 --- Clinical UI Layer}

The clinical UI layer presents diagnostic outputs to users through two
distinct interaction modes, each designed for a specific user context
and subject to appropriate clinical safeguards.

\textbf{Clinician mode} is designed for use by licensed psychiatric
clinicians as a decision support aid during or after clinical
consultations. In this mode, the platform presents the full ranked
differential diagnosis with DSM-5 code mappings, per-criterion symptom
evidence, confidence indicators for each candidate diagnosis, and
explanatory output derived from the ensemble reasoning process.
Clinicians retain full control over interpretation and final diagnostic
determination; the platform's outputs are explicitly framed as
AI-generated decision support, not clinical diagnoses.

\textbf{Patient mode} provides a guided self-screening interface
for patients or service members seeking to assess their own mental
health status. This mode employs a structured conversational intake
flow aligned with validated screening instruments~\cite{Kroenke2001,
Weathers2013} and returns structured self-assessment feedback rather
than raw diagnostic labels. Escalation prompts are embedded throughout
the intake flow to detect indicators of acute risk --- including
expressions of suicidal ideation or severe functional impairment ---
and direct users to immediate human support resources.

\subsection{Privacy and Security Architecture}

The zero-egress guarantee is enforced through a combination of
architectural constraints, runtime controls, and local data management
practices. Figure~\ref{fig:architecture} illustrates the zero-egress
boundary that encloses all system components.

\textbf{Data flow isolation.} All inference, orchestration, and storage
operations execute within the mobile device's application sandbox. No
network sockets are opened during inference. Patient conversational data
is processed exclusively in volatile memory during active sessions and
is never written to persistent storage without explicit, user-authorized
action.

\textbf{Local encryption.} Where session data is retained at user
request (e.g., for clinician-reviewed record-keeping), it is encrypted
using AES-256-GCM with device-bound keys managed by the platform's
secure enclave~\cite{ARMTrustZone, AppleSecureEnclave}. Encryption keys
are never exported from the device.

\textbf{No telemetry or analytics.} The platform contains no telemetry
collection, crash reporting, usage analytics, or model update mechanisms
that would result in any data transmission. Model weights are fixed at
deployment and are updated only through explicit, user-initiated
application updates distributed via standard platform channels.

\textbf{Session isolation.} Each clinical session is isolated at the
application level. No data from a previous session is accessible to a
subsequent session unless explicitly exported and re-imported by an
authorized user. This design prevents inference attacks that could
reconstruct patient identity from residual session data.

\textbf{Threat model.} The proposed architecture protects against the
following threats: unauthorized access to patient conversational data
by third-party cloud providers; data interception during transmission;
re-identification of patients from telemetry or usage logs; and
unauthorized model updates that could introduce adversarial behavior.
It does not protect against physical device compromise by an adversary
with direct hardware access, which is outside the scope of the software
architecture and is addressed at the device operating system level.


\section{On-Device LLM Deployment}
\label{sec:deployment}

\begin{figure}[H]
\centering
\includegraphics[width=4.8in]{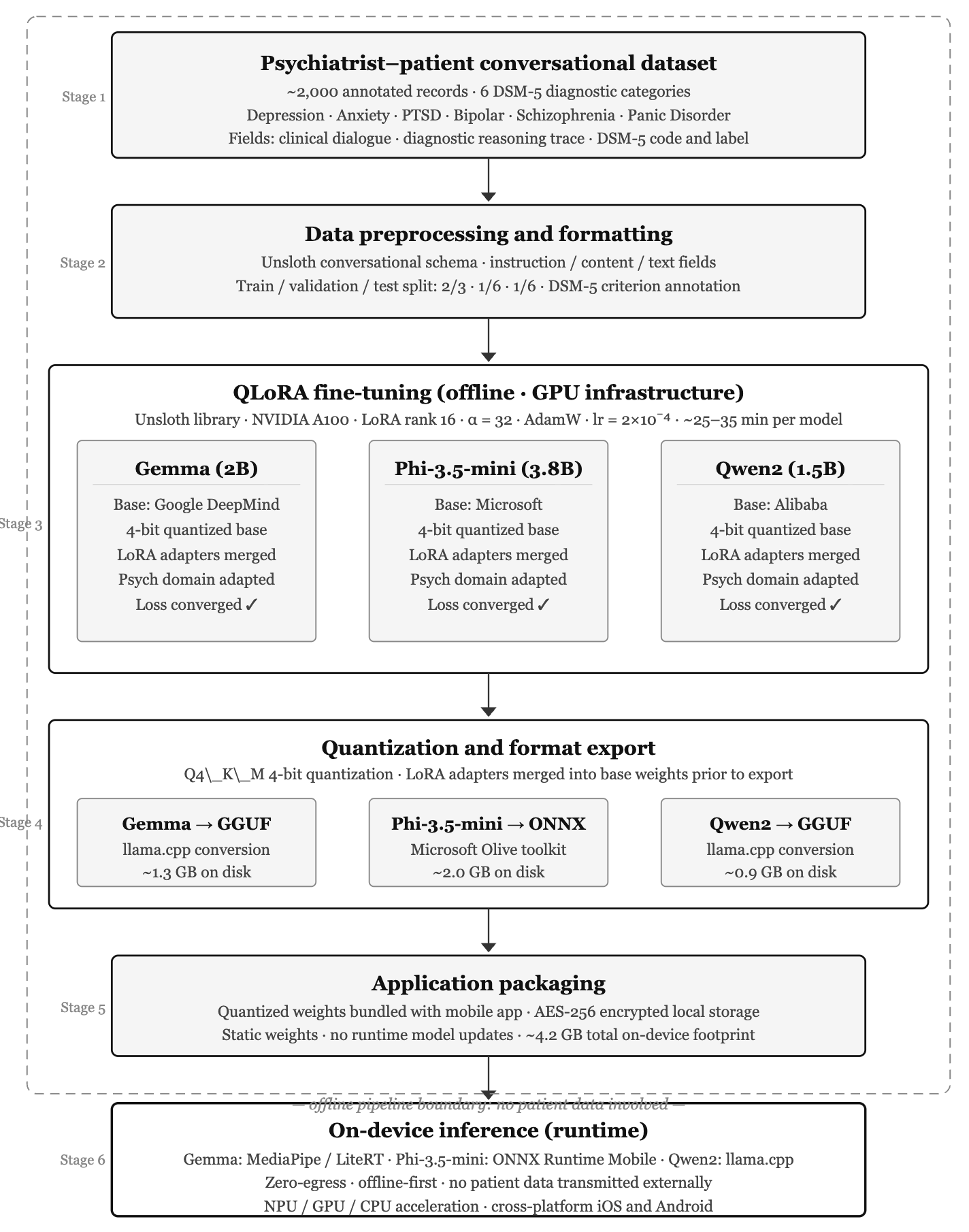}
\DeclareGraphicsExtensions.
\caption{End-to-end offline fine-tuning and on-device deployment
pipeline. The \texttt{mental-reasoning} dataset is preprocessed into
a DSM-5 aligned instruction format and used to fine-tune three
lightweight open-source LLMs (Gemma, Phi-3.5-mini, and Qwen2) using
QLoRA on GPU infrastructure. Fine-tuned models are quantized to 4-bit
precision and exported to mobile-compatible formats (GGUF for Gemma
and Qwen2, ONNX for Phi-3.5-mini). Quantized model weights are
packaged with the mobile application and executed at runtime via
platform-native inference runtimes entirely on the user's device.
No patient data is involved in any stage of this pipeline beyond the
device boundary at inference time.}
\label{fig:deployment_pipeline}
\end{figure}

This section details the end-to-end pipeline through which
general-purpose open-source language models are transformed into
privacy-preserving, on-device psychiatric decision support tools.
The pipeline comprises four stages: offline fine-tuning on
domain-specific clinical data, model quantization and export to
mobile-compatible formats, runtime deployment via mobile inference
frameworks, and on-device ensemble orchestration. Preliminary
performance observations are reported in Section~\ref{subsec:perf}.
Figure~\ref{fig:deployment_pipeline} illustrates the complete
pipeline from training data to on-device inference.

\subsection{Fine-Tuning Dataset}
\label{subsec:dataset}

Fine-tuning is performed on the \texttt{mental-reasoning} dataset,
a purpose-built collection of psychiatrist--patient clinical
conversations publicly released by the authors on
HuggingFace~\cite{mentalreasoning2025}. The dataset comprises 500
annotated conversational records in Parquet format, each structured
as a four-field schema: an \texttt{instruction} field containing a
fixed system prompt establishing the model's role as a psychiatric
diagnostic assistant; a \texttt{conversation} field containing a
multi-turn psychiatrist--patient dialogue (ranging from approximately
2,070 to 5,070 characters); a \texttt{diagnosis} field containing
the DSM-5 disorder label and code; and a \texttt{condition} field
containing the short-form diagnostic category label.

The dataset spans five primary psychiatric diagnostic categories
aligned with DSM-5 criteria~\cite{APA2013}: Major Depressive
Disorder, Generalized Anxiety Disorder, Post-Traumatic Stress
Disorder~(PTSD), Bipolar Disorder, and Schizophrenia. Each
conversational record simulates a realistic clinical encounter in
which the clinician systematically elicits symptom history,
duration, severity, and functional impact aligned with DSM-5
diagnostic thresholds. Diagnoses are expressed using standardized
DSM-5 codes~(e.g., Major Depressive Disorder: DSM-5 296.2x).

The dataset was partitioned into training, validation, and test
subsets using a 2/3, 1/6, 1/6 split, yielding approximately 333
training, 83 validation, and 84 test records respectively.

\subsection{Fine-Tuning Pipeline}
\label{subsec:finetuning}

Fine-tuning is conducted entirely offline, prior to application
deployment, on GPU infrastructure. The resulting model weights are
static at runtime --- no online learning or model updates occur on
the device.

\subsubsection{Data Preprocessing and Formatting}

The Unsloth fine-tuning library~\cite{unsloth2024} requires training
data to be structured in a standardized conversational schema.
Each record from the \texttt{mental-reasoning} dataset maps directly
onto the required three-field Unsloth format: the \texttt{instruction}
field provides the task context and diagnostic framing; the
\texttt{conversation} field provides the clinical dialogue as the
primary input; and the \texttt{diagnosis} field provides the expected
model output as the DSM-5-aligned diagnostic label and code. This
schema is applied uniformly across all three models, ensuring
consistency in fine-tuning objectives and output format.

\subsubsection{Training Configuration}

Fine-tuning is conducted using the Unsloth
library~\cite{unsloth2024} on NVIDIA A100 GPU infrastructure via
Google Colab, leveraging Low-Rank Adaptation~(LoRA)~\cite{hu2021lora}
with 4-bit base model quantization
(QLoRA)~\cite{dettmers2023qlora}. The LoRA configuration uses a
rank of $r = 16$, scaling factor $\alpha = 32$, and dropout of
$0.05$, applied to the query, key, value, and output projection
matrices of all self-attention layers. The AdamW optimizer is used
with a learning rate of $2 \times 10^{-4}$ and a cosine annealing
schedule. Each model is trained for a maximum of 3 epochs with
early stopping based on validation loss, using a batch size of 4
with gradient accumulation over 4 steps (effective batch size of
16). Peak GPU memory utilization during training reached
approximately 14.6~GB. Training completion times per model ranged
from approximately 25 to 35 minutes on this hardware
configuration. 


\subsection{Model Quantization and Export}
\label{subsec:quantization}

Following fine-tuning, each model undergoes quantization and format
conversion to produce weights suitable for efficient on-device
inference. LoRA adapter weights are merged into the base model
prior to quantization to produce a single unified weight file per
model.

The Q4\_K\_M quantization scheme is applied for GGUF
models~\cite{gguf2023}, which retains higher precision for
attention and embedding layers while applying 4-bit quantization
to the majority of weight tensors, balancing compression ratio
with inference quality. Gemma and Qwen2 are exported to GGUF
format using \texttt{llama.cpp} conversion
utilities~\cite{llamacpp2023, nurolense}. Phi-3.5-mini is exported to ONNX
format using Microsoft's Olive optimization
toolkit~\cite{abdin2024phi3}, applying INT4 weight quantization
and graph optimizations tailored for ONNX Runtime Mobile.

Table~\ref{tab:model_sizes} summarizes the resulting model sizes
and on-device memory requirements after quantization.

\begin{table*}[!htb]
\centering
\caption{Model sizes and on-device memory requirements after QLoRA 4-bit quantization.}
\begin{adjustbox}{width=1.00\textwidth}
\label{tab:model_sizes}
\begin{tabular}{lcccc}
\toprule
\thead{Model} & \thead{Parameters} & \thead{Format} & \thead{Disk size} & \thead{Runtime memory} \\
\midrule
Gemma (Fast)         & 2B   & GGUF Q4\_K\_M & $\sim$0.53~GB & $\sim$0.7~GB \\
Gemma (Full)         & 2B   & GGUF Q4\_K\_M & $\sim$1.5~GB  & $\sim$1.7~GB \\
Phi-3.5-mini         & 3.8B & ONNX INT4     & $\sim$2.0~GB  & $\sim$2.2~GB \\
Qwen2                & 1.5B & GGUF Q4\_K\_M & $\sim$0.9~GB  & $\sim$1.1~GB \\
\midrule
\textbf{Total ensemble} & --- & ---          & $\sim$4.9~GB  & $\sim$5.7~GB \\
\bottomrule
\end{tabular}
\end{adjustbox}
\end{table*}

\subsection{Mobile Inference Runtime}
\label{subsec:runtime}

Each model is executed at runtime via a platform-native inference
framework selected for compatibility with the model's export
format, mobile hardware acceleration support, and cross-platform
portability.

\textbf{Gemma --- MediaPipe / LiteRT.}
Gemma is executed via Google's MediaPipe LLM Inference
API~\cite{team2024gemma}, built on the LiteRT runtime. Two
quantization variants are supported: a lightweight Fast variant
($\sim$529~MB) optimized for low-latency response on
resource-constrained devices, and a Full variant ($\sim$1.5~GB)
offering higher output quality. The runtime supports NPU
acceleration on Qualcomm Hexagon, Apple Neural Engine, and Google
Tensor processing units.

\textbf{Phi-3.5-mini --- ONNX Runtime Mobile.}
Phi-3.5-mini is executed via ONNX Runtime
Mobile~\cite{abdin2024phi3}, supporting hardware acceleration via
NNAPI on Android and Core ML on iOS. The INT4 ONNX model produced
by the Olive pipeline is directly consumed by the runtime, with
automatic selection of the optimal acceleration backend on the
target device.

\textbf{Qwen2 --- llama.cpp.}
Qwen2 is executed via llama.cpp~\cite{llamacpp2023}, a portable
C/C++ inference engine supporting GGUF models across Android
(OpenCL / Vulkan) and iOS (Metal Performance Shaders). Thread
count and context window size are configured dynamically at runtime
based on available device memory, ensuring stable operation across
mid-range and flagship hardware.

\subsection{On-Device Ensemble Orchestration}
\label{subsec:orchestration}

\begin{figure}[H]
\centering
\includegraphics[width=4.8in]{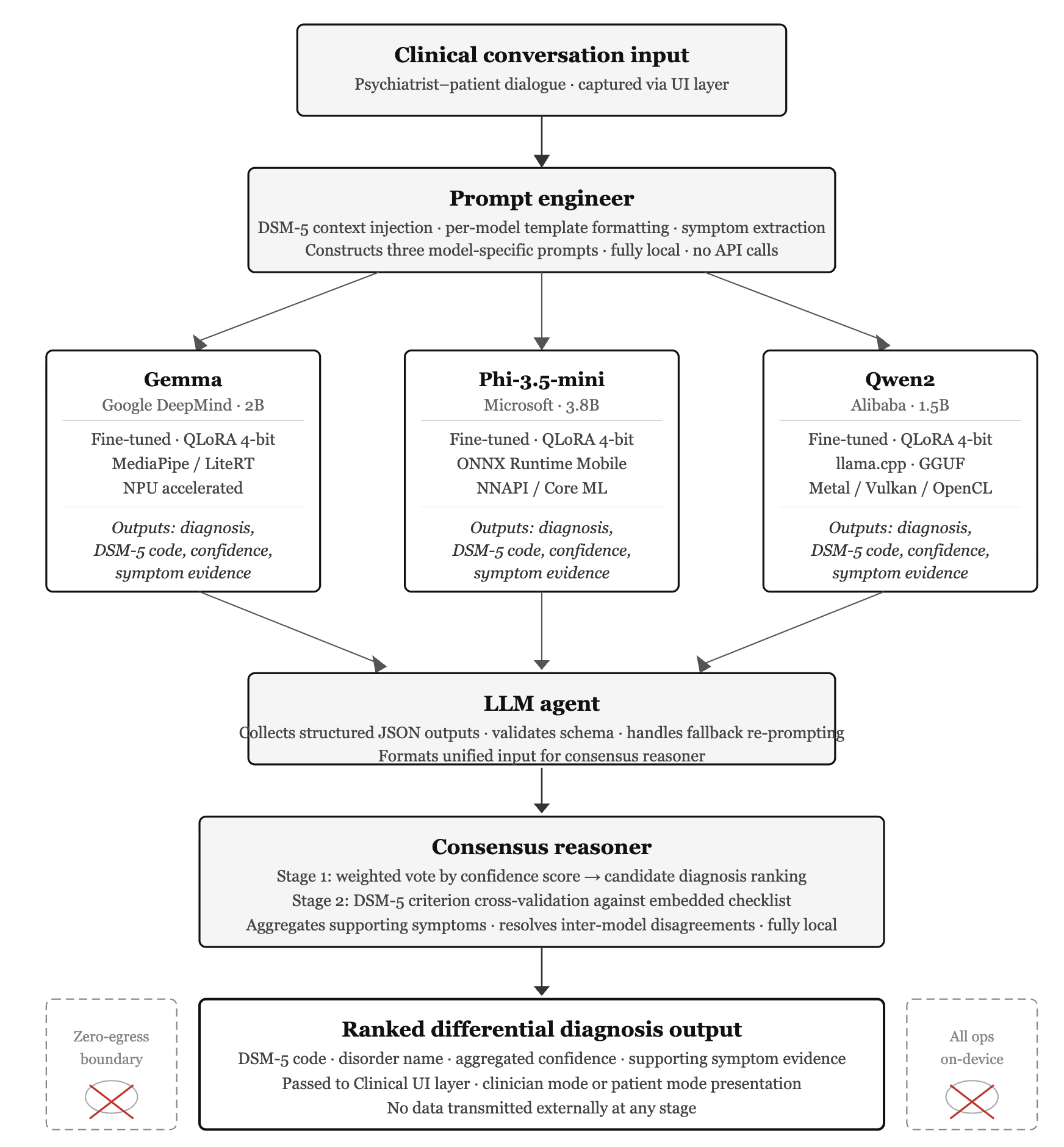}
\DeclareGraphicsExtensions.
\caption{On-device ensemble inference flow. A clinical conversation
is captured via the UI layer and passed to the prompt engineer,
which constructs model-specific prompts embedding DSM-5 diagnostic
context. Each of the three fine-tuned models (Gemma, Phi-3.5-mini,
Qwen2) is invoked by the LLM agent, producing structured diagnostic
outputs comprising a candidate diagnosis, DSM-5 code, confidence
score, and supporting symptom evidence. The consensus reasoner
aggregates these outputs via weighted voting and DSM-5
criterion cross-validation, producing a ranked differential
diagnosis list returned to the UI layer. All operations execute
entirely on-device.}
\label{fig:inference_flow}
\end{figure}

At runtime the three models are coordinated by the on-device
orchestration layer. Figure~\ref{fig:inference_flow} illustrates
the end-to-end inference flow.

\subsubsection{Prompt Construction}

Each model is prompted using a fixed system instruction drawn
directly from the \texttt{mental-reasoning} dataset schema:
\textit{``You are a psychiatric diagnostic assistant. Analyze the
following psychiatrist--patient conversation and provide the DSM-5
diagnosis.''} This instruction is prepended to the clinical
conversation transcript, with DSM-5 criterion context injected as
a structured reference block. Prompt templates are tailored to each
model's training format. All prompt construction uses static
templates stored within the application bundle; no external prompt
retrieval or update calls are made at runtime.

\subsubsection{Output Schema}

Each model is prompted to produce a structured JSON output:

\begin{verbatim}
{
  "diagnosis": "<DSM-5 disorder name>",
  "dsm5_code": "<DSM-5 numeric code>",
  "confidence": <float 0.0 -- 1.0>,
  "supporting_symptoms": ["<symptom_1>", "<symptom_2>", ...],
  "differential": [{"diagnosis": "...", "confidence": ...}]
}
\end{verbatim}

Outputs that do not conform to this schema are flagged by the LLM
agent, which re-prompts the offending model up to two times before
treating its output as unavailable for that inference cycle.

\subsubsection{Consensus Algorithm}

The consensus reasoner aggregates model outputs in two stages.
In stage one, candidate diagnoses are grouped by DSM-5 code and
a weighted vote count computed, with each model's vote weighted
by its reported confidence score. In stage two, the top-ranked
candidate is cross-validated against the DSM-5 criterion checklist
embedded in the application, verifying that supporting symptoms
cited by at least two of the three models satisfy the minimum
diagnostic threshold~\cite{APA2013}. The final output is a ranked
differential diagnosis list with aggregated confidence scores,
symptom evidence, and DSM-5 code mappings.

\subsection{Preliminary Performance Observations}
\label{subsec:perf}

Preliminary response-time observations were obtained through
visual inspection of a demonstration session of the mobile
application. Response times were estimated using
Time-to-First-Visible-Response~(TTFVR), defined as the elapsed
time between prompt submission~($T_0$) and the appearance of the
first visible output token in the response area~($T_1$), such
that $\text{TTFVR} = T_1 - T_0$.

These observations should be interpreted cautiously. They are
demo-observed estimates derived from a single uncontrolled session
rather than from an instrumented benchmark protocol, and are
subject to an estimated measurement uncertainty of approximately
$\pm$0.5 to 1.0 seconds arising from screen-recording frame rate,
UI animation transitions, and video compression artefacts.
Furthermore, the hardware specification of the test device is not
documented, cold-start versus warm-inference state is unknown,
and cloud response timing may have been influenced by transient
network conditions. These figures are therefore presented as
preliminary indicative estimates only; a controlled benchmark study
is required before publication-quality comparative performance
claims can be made.

Table~\ref{tab:perf} summarises the preliminary observations.

\begin{table*}[!htb]
\centering
\caption{Preliminary demo-observed response-time estimates.
TTFVR = Time-to-First-Visible-Response. All values are
single-session visual estimates; see text for limitations.}
\begin{adjustbox}{width=1.00\textwidth}
\label{tab:perf}
\begin{tabular}{lccc}
\toprule
\thead{Mode} & \thead{Model size} & \thead{TTFVR (s)} & \thead{Evidence status} \\
\midrule
Gemma Fast (on-device) & $\sim$529~MB & 7.5--8.0     & Observable in demo video \\
Gemma Full (on-device) & $\sim$1.5~GB & Not measured & \makecell{Confirmed available\\No completed run visible} \\
Cloud AI (baseline)    & N/A          & 2.5--3.0     & Observable reference baseline \\
\bottomrule
\end{tabular}
\end{adjustbox}
\end{table*}

The only on-device configuration for which a defensible latency
estimate can be established from the current evidence is the
Gemma Fast variant ($\sim$529~MB), which produced a TTFVR of
approximately 7.5 to 8.0 seconds. The cloud-backed baseline
exhibited materially faster first-response times of approximately
2.5 to 3.0 seconds under comparable task conditions, consistent
with the latency advantages of server-side GPU inference over
on-device CPU/NPU execution. The Gemma Full model ($\sim$1.5~GB)
is confirmed as present and downloadable within the application,
but its inference latency remains undetermined from the current
evidence.

The observed on-device latency is within a clinically acceptable
range for a decision support tool used during or after a
consultation, where response times of under 10 seconds are
generally tolerable~\cite{laskaridis2024}. However, the latency
gap between on-device and cloud inference underscores the
inherent tradeoff between privacy preservation and response speed
that motivates ongoing optimization work described in
Section~\ref{sec:conclusion}.

A controlled benchmark protocol is recommended for future
evaluation, comprising a fixed test device with documented
hardware specification, a standardised clinical prompt corpus of
at least five representative prompts per diagnostic category,
five to ten repeated runs per model per prompt in airplane mode
for on-device conditions and stable network conditions for cloud
baseline runs, and reporting of median and interquartile range
TTFVR, token throughput, total completion time, and peak memory
footprint.


\section{Clinical Functionality}
\label{sec:functionality}

This section describes the clinical capabilities of the proposed
platform as implemented in the mobile application. The platform
operates as a conversational clinical decision support assistant,
exposing a set of structured clinical task flows through a
natural language interface accessible to clinicians at the point
of care. All functionality is available in both cloud-backed and
fully on-device privacy-preserving modes, selectable by the
clinician based on the sensitivity of the clinical context.

\subsection{AI Model Selection and Privacy Modes}
\label{subsec:modes}

A central design feature of the platform is its explicit, user-visible
AI model selection mechanism. At application startup and through the
Settings panel, clinicians are presented with three clearly differentiated
operating modes, as shown in Figure~\ref{fig:settings}:

\begin{figure}[H]
\centering
\includegraphics[width=2.6in]{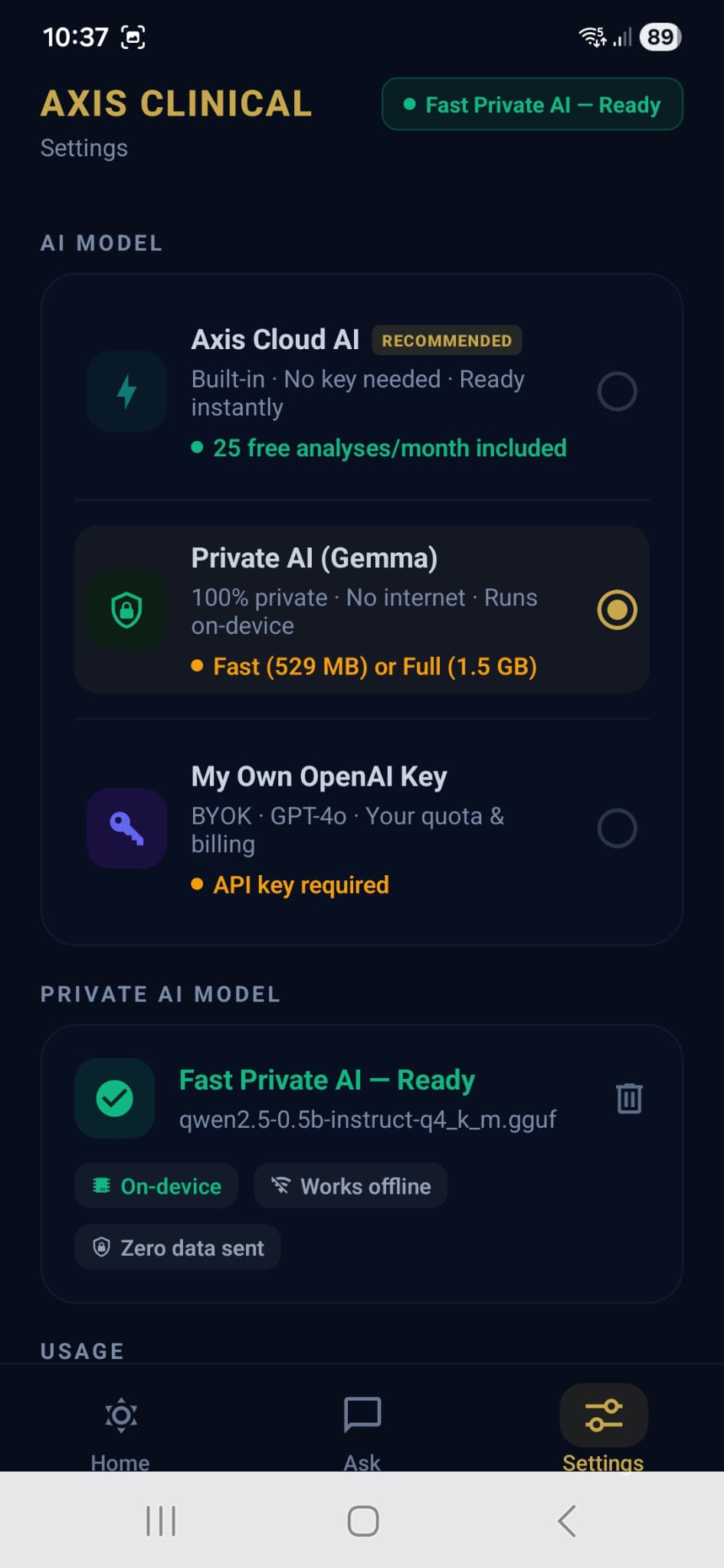}
\DeclareGraphicsExtensions.
\caption{AI model selection panel in the mobile application.
Three modes are available: (1)~Cloud AI, a server-backed mode
offering fast response with 25 free analyses per month included;
(2)~Private AI, a fully on-device mode labeled
``100\% private $\cdot$ No internet $\cdot$ Runs on-device'',
which internally deploys a fine-tuned LLM ensemble for
consensus-based psychiatric reasoning; and (3)~Bring Your Own
Key (BYOK), enabling integration with a user-supplied OpenAI
API key. The settings panel displays three explicit privacy
guarantees for the Private AI mode: \textit{On-device},
\textit{Works offline}, and \textit{Zero data sent}. The
currently active on-device model file is shown in the Private
AI Model section, confirming the exact quantized weight in use.}
\label{fig:settings}
\end{figure}

\textbf{Cloud AI mode} routes inference to a server-backed endpoint
and is recommended for non-sensitive general clinical tasks where
response speed is prioritized. This mode includes 25 free analyses
per month and requires no API configuration.

\textbf{Private AI mode} is the privacy-preserving configuration
recommended for all psychiatric applications. In this mode, the
platform deploys a consortium of three fine-tuned, quantized
LLMs --- Gemma, Phi-3.5-mini, and Qwen2 --- entirely on the
device, with no data transmitted externally at any stage. While
the current application interface exposes Gemma as the primary
visible on-device model (available in Fast $\sim$529~MB and Full
$\sim$1.5~GB variants), the underlying psychiatric decision
support pipeline invokes all three fine-tuned models in parallel
via the on-device orchestration layer described in
Section~\ref{sec:deployment}. The settings panel confirms the
active quantized model file in use
(e.g.,~\texttt{qwen2.5-0.5b-instruct-q4\_k\_m.gguf}), providing
clinician-visible transparency over which model weights are
executing on the device. The Private AI mode is explicitly
labeled with three guarantees: \textit{On-device},
\textit{Works offline}, and \textit{Zero data sent} ---
enforced at the architecture level, not through policy.

The multi-model ensemble design is a deliberate clinical safety
choice. Rather than relying on a single model's output for
psychiatric assessment --- which carries the risk of systematic
errors or overconfident predictions --- the platform invokes all
three fine-tuned models independently for each query and
aggregates their outputs through a consensus-based reasoning
process~\cite{bandara2025standardization}. Disagreements between
models are surfaced as low-confidence flags to the clinician,
and only diagnoses supported by at least two of the three models
and validated against DSM-5 criteria are promoted to the primary
output. This consensus-driven approach meaningfully improves
diagnostic robustness and reduces the risk of single-model
failure modes in high-stakes psychiatric contexts, while
preserving the zero-egress guarantee throughout.

\textbf{Bring Your Own Key (BYOK) mode} allows clinicians to
supply their own OpenAI API key, enabling GPT-4o access under
the clinician's own quota and billing arrangements. This mode
is intended for institutions with existing OpenAI enterprise
agreements and does not invoke the on-device ensemble.

For psychiatric applications --- the primary focus of this paper
--- Private AI mode with the full on-device LLM ensemble is the
recommended configuration, ensuring that sensitive patient
conversational data never leaves the device regardless of network
availability, while delivering consensus-driven diagnostic
reasoning across all supported conditions.

\subsection{Clinical Task Flows}
\label{subsec:tasks}

\begin{figure}[H]
\centering
\includegraphics[width=2.6in]{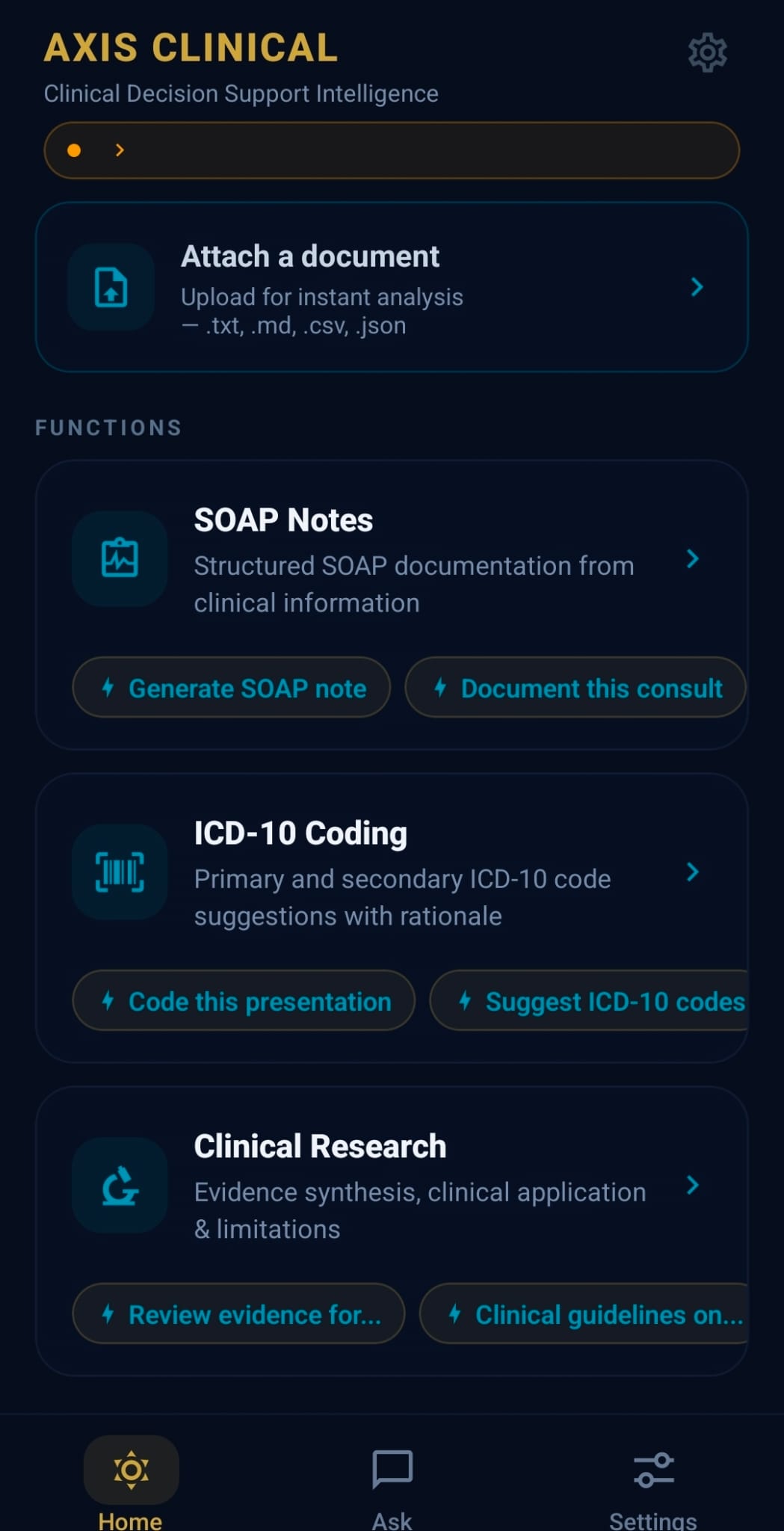}
\DeclareGraphicsExtensions.
\caption{Home screen of the mobile application, subtitled
\textit{Clinical Decision Support Intelligence}. The screen
presents a document attachment entry point supporting
.txt, .md, .csv, and .json formats, followed by three
primary clinical function cards under a \textsc{Functions}
heading: (1)~\textbf{SOAP Notes} --- \textit{Structured SOAP
documentation from clinical information} --- with quick-action
buttons \textit{Generate SOAP note} and \textit{Document this
consult}; (2)~\textbf{ICD-10 Coding} --- \textit{Primary and
secondary ICD-10 code suggestions with rationale} --- with
quick-action buttons \textit{Code this presentation} and
\textit{Suggest ICD-10 codes}; and (3)~\textbf{Clinical
Research} --- \textit{Evidence synthesis, clinical application
\& limitations} --- with quick-action buttons \textit{Review
evidence for\ldots} and \textit{Clinical guidelines on\ldots}.
Bottom navigation provides Home, Ask, and Settings tabs.}
\label{fig:home}
\end{figure}

The platform exposes four primary clinical task flows through
its home screen, each accessible via a dedicated function card
with one-tap quick-action buttons, as shown in
Figure~\ref{fig:home}.

When operating in Private AI mode, all diagnostic and clinical
task flows are powered by the on-device fine-tuned LLM ensemble
described in Section~\ref{sec:deployment}. Clinician queries ---
including differential diagnosis requests, symptom assessments,
ICD-10 coding, and SOAP note generation --- are processed
entirely by the Gemma, Phi-3.5-mini, and Qwen2 ensemble via
the on-device orchestration layer, without any external inference
call. Each query is routed through the prompt engineer, which
constructs a model-specific prompt embedding the clinical input
and DSM-5 context, before being dispatched to the three fine-tuned
models. The consensus reasoner then aggregates the ensemble
outputs into a final ranked response returned to the UI layer.
This end-to-end on-device inference pipeline means that for
psychiatric applications in particular --- where patient
conversational content is maximally sensitive --- the fine-tuned
diagnostic capability of the ensemble is delivered with a
zero-egress guarantee at every step.

\subsubsection{SOAP Notes}

The SOAP Notes function generates structured clinical documentation
in the standard Subjective, Objective, Assessment, and Plan format
from clinician-provided patient information. As shown in
Figure~\ref{fig:soap}, upon triggering the \textit{Generate SOAP
note} function the platform returns a structured intake prompt
requesting four categories of clinical input:
\textit{Subjective}~(patient complaints, symptoms, and history);
\textit{Objective}~(vital signs, physical examination findings,
lab results, and imaging); \textit{Assessment}~(clinical
impressions and differential diagnoses); and
\textit{Plan}~(proposed treatment, further investigations,
referrals, or follow-up). The platform processes this structured
input and generates a formatted SOAP note suitable for inclusion
in the patient record. Two quick-action variants are available:
\textit{Generate SOAP note} (from clinician-provided details)
and \textit{Document this consult} (from a free-form
consultation summary). A model attribution label is displayed
below each AI response --- showing ``Cloud AI'' or ``Private AI''
--- providing continuous transparency to the clinician about
which inference mode produced the output. When operating in
Private AI mode, SOAP note generation executes entirely
on-device, ensuring that no patient identifiers or clinical
content are transmitted externally~\cite{hipaa1996}.

\subsubsection{ICD-10 Coding}

The ICD-10 Coding function provides primary and secondary
International Classification of Diseases, Tenth Revision~(ICD-10)
code suggestions with supporting clinical rationale derived from
the presented case information~\cite{WHO2019ICD}. The clinician can
either submit a free-text case description or attach a structured
clinical document for analysis. The platform returns a ranked list
of ICD-10 code suggestions with brief rationale for each, supporting
accurate and consistent clinical coding at the point of care.
Quick-action prompts include \textit{Code this presentation} and
\textit{Suggest ICD-10 codes}.

\subsubsection{Clinical Research}

The Clinical Research function provides evidence synthesis,
clinical guideline summarization, and literature-based decision
support. Clinicians can query the platform for summaries of
clinical guidelines, evidence reviews for specific interventions,
or drug interaction assessments. Suggested quick-action prompts
visible in the application include \textit{Summarise the latest
NICE guidelines} and \textit{Review evidence for...}, reflecting
the platform's integration with current evidence-based practice
frameworks.

\subsubsection{Document Attachment and Analysis}

The platform supports attachment of clinical documents in plain
text, Markdown, CSV, and JSON formats for instant AI-assisted
analysis. Clinicians can attach clinical notes, referral letters,
patient-reported outcome measures, or structured data files and
pose natural language questions about their content. This enables
use cases such as reviewing a referral letter for relevant
psychiatric history, extracting structured symptom data from
unstructured clinical notes, or summarizing a longitudinal patient
record prior to a consultation.

\subsection{Conversational Interface}
\label{subsec:interface}

\begin{figure}[H]
\centering
\includegraphics[width=2.6in]{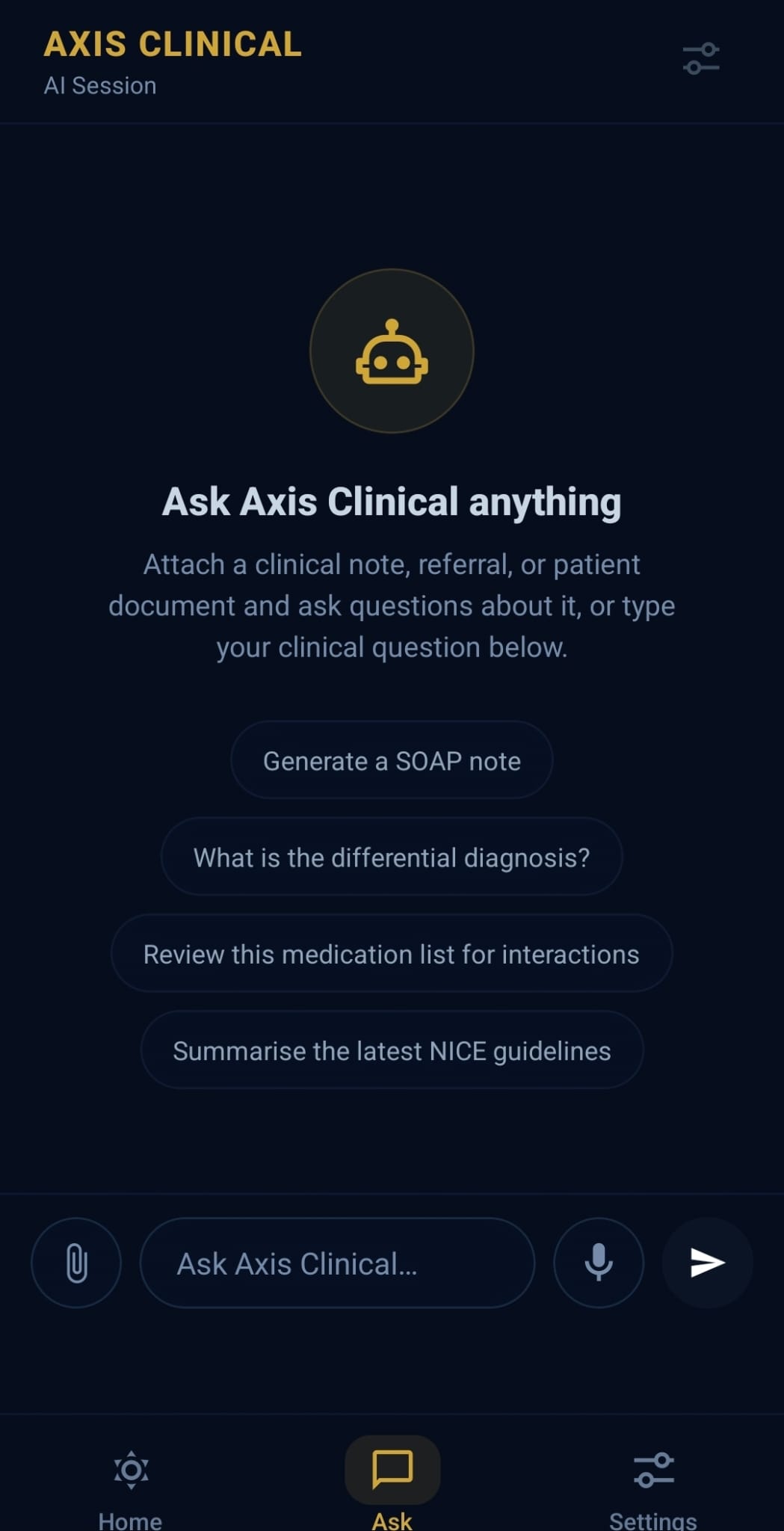}
\DeclareGraphicsExtensions.
\caption{Conversational interface (AI Session screen) of the
mobile application. The interface presents four contextual
suggested prompt chips: \textit{Generate a SOAP note},
\textit{What is the differential diagnosis?}, \textit{Review
this medication list for interactions}, and \textit{Summarise
the latest NICE guidelines}. The bottom input bar provides
typed entry, voice input via microphone, document attachment,
and a send button. Bottom navigation tabs (Home, Ask, Settings)
are persistent across all screens.}
\label{fig:ask}
\end{figure}

All clinical task flows are accessed through a unified
conversational interface, as shown in Figure~\ref{fig:ask}.
The interface presents a natural language input field with
support for both typed and voice input, an attachment button
for document upload, and a set of suggested prompt chips that
surface the most clinically relevant queries for the active
task context.

\begin{figure}[H]
\centering
\includegraphics[width=2.6in]{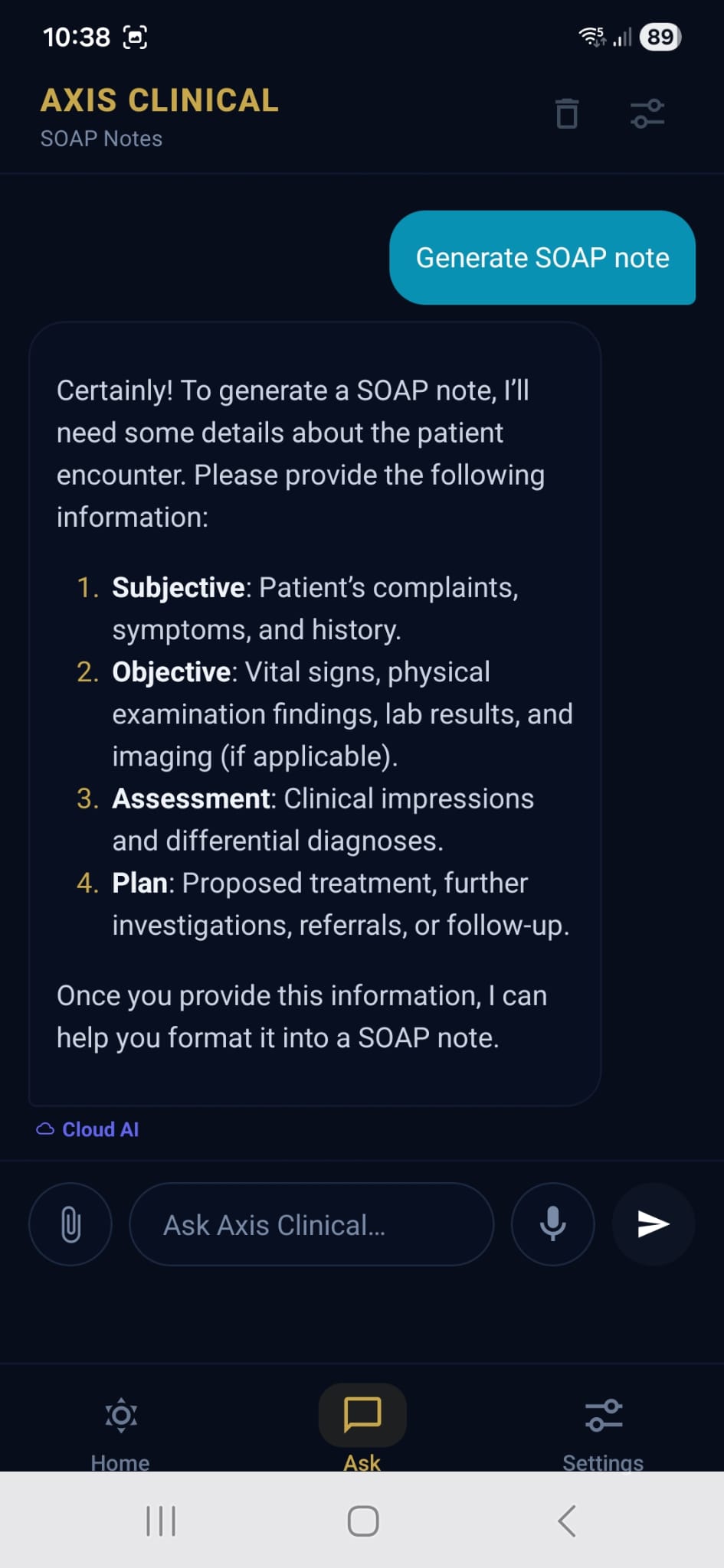}
\DeclareGraphicsExtensions.
\caption{SOAP Notes task flow. Upon receiving a
\textit{Generate SOAP note} request, the platform returns a
structured intake prompt requesting four categories of clinical
information: (1)~\textbf{Subjective} --- patient complaints,
symptoms, and history; (2)~\textbf{Objective} --- vital signs,
physical examination findings, lab results, and imaging;
(3)~\textbf{Assessment} --- clinical impressions and
differential diagnoses; and (4)~\textbf{Plan} --- proposed
treatment, further investigations, referrals, or follow-up.
The AI model attribution label (``Cloud AI'') is displayed
below the response, providing continuous transparency about
which inference mode produced the output. When operating in
Private AI mode, this label reads ``Private AI'' confirming
on-device execution.}
\label{fig:soap}
\end{figure}

Figure~\ref{fig:soap} illustrates the SOAP Notes task flow,
showing the structured intake prompt the platform returns
upon receiving a \textit{Generate SOAP note} request.

The conversational design reflects a clinician-in-the-loop
interaction model in which the platform provides structured
suggestions and evidence-linked outputs but does not autonomously
act or make definitive clinical determinations. All outputs are
framed as AI-assisted decision support, and clinicians retain full
interpretive authority over the information presented.

\subsection{Supported Psychiatric Conditions}
\label{subsec:conditions}

For psychiatric decision support specifically, the platform's
on-device fine-tuned LLM ensemble is trained to recognize and
assess five primary diagnostic categories from the
\texttt{mental-reasoning} dataset~\cite{mentalreasoning2025},
all aligned with DSM-5 diagnostic criteria~\cite{APA2013}.
Table~\ref{tab:conditions} summarizes the supported conditions,
their DSM-5 codes, primary symptom domains assessed, and
associated validated screening instruments.

\begin{table}[H]
\centering
\caption{Psychiatric conditions supported by the on-device
fine-tuned LLM ensemble, with DSM-5 codes, primary symptom
domains, and associated validated screening instruments.}
\label{tab:conditions}
\renewcommand{\arraystretch}{1.4}
\begin{tabular}{p{3.2cm} p{1.8cm} p{4.0cm} p{3.2cm}}
\hline
\textbf{Condition} & \textbf{DSM-5 code} &
\textbf{Primary symptom domains} &
\textbf{Screening instrument} \\
\hline
Major Depressive Disorder
& 296.2x
& Depressed mood, anhedonia, sleep/appetite disturbance, fatigue, concentration, psychomotor change, suicidality
& PHQ-9~\cite{Kroenke2001} \\
Generalized Anxiety Disorder
& 300.02
& Excessive worry, restlessness, fatigue, concentration difficulty, irritability, muscle tension, sleep disturbance
& GAD-7~\cite{Spitzer2006} \\
Post-Traumatic Stress Disorder
& 309.81
& Intrusion symptoms, avoidance, negative cognition/mood, hyperarousal, trauma exposure history
& PCL-5~\cite{Weathers2013} \\
Bipolar Disorder
& 296.4x--296.8x
& Manic/hypomanic episodes, elevated mood, grandiosity, decreased sleep, impulsivity, depressive episodes
& MDQ~\cite{Hirschfeld2000} \\
Schizophrenia
& 295.90
& Positive symptoms (hallucinations, delusions, disorganized speech), negative symptoms, functional decline
& PANSS~\cite{Kay1987} \\
\hline
\end{tabular}
\end{table}

\subsection{Clinical Safeguards and Responsible AI Design}
\label{subsec:safeguards}

The platform incorporates several layers of safeguards to ensure
responsible clinical use, consistent with emerging frameworks for
AI in healthcare~\cite{Topol2019, Chen2021bias}.

\textbf{Decision support framing.} All diagnostic outputs are
explicitly framed as AI-assisted decision support rather than
clinical diagnoses. The platform does not claim diagnostic
authority, and all outputs are accompanied by language that
reinforces the clinician's interpretive responsibility.

\textbf{Confidence communication.} The ensemble consensus output
includes per-candidate confidence scores and a ranked differential
diagnosis list, enabling clinicians to assess the degree of model
certainty and consider alternative diagnoses. Cases with low
ensemble consensus --- where model outputs diverge substantially ---
are flagged to the clinician as low-confidence outputs requiring
independent clinical assessment.

\textbf{Escalation protocol.} Conversational inputs containing
indicators of acute psychiatric risk --- including expressions of
suicidal ideation, self-harm intent, or severe functional
impairment --- trigger an escalation prompt that directs the
clinician or patient to immediate human support resources,
consistent with safe messaging guidelines~\cite{Suicide2022}.

\textbf{Transparency of AI mode.} Every AI-generated response
in the application displays a model attribution label immediately
below the output --- ``Cloud AI'' when inference is server-backed,
or ``Private AI'' when executing on-device --- as visible in
Figure~\ref{fig:soap}. This persistent per-response labeling
ensures clinicians are always aware of which inference mode
produced a given output, supporting informed consent and
institutional data governance compliance.

\textbf{Bias acknowledgment and dataset limitations.} The
fine-tuning dataset~\cite{mentalreasoning2025} comprises 500
records spanning five diagnostic categories. While this dataset
provides a strong foundation for proof-of-concept evaluation,
its demographic and linguistic representativeness has not been
fully characterized. Outputs may reflect biases present in the
training data, including potential under-representation of
certain demographic groups, cultural variation in symptom
expression, and diagnostic assumptions embedded in the training
annotations. These limitations are acknowledged in the platform
documentation and inform the ongoing data collection efforts
described in Section~\ref{sec:conclusion}.


\section{Related Work}
\label{sec:related}

This section surveys the body of work most directly relevant to
the proposed platform, organized across four areas: LLMs for
psychiatric and mental health applications; privacy-preserving
AI in mental healthcare; mobile and edge AI for clinical
decision support; and existing mHealth platforms for psychiatric
care. Table~\ref{tab:related} provides a structured comparison
of representative works against the proposed platform across
key dimensions.

\subsection{LLMs for Psychiatric and Mental Health Applications}

The application of LLMs to psychiatric and
mental health tasks has expanded substantially since 2023.
Early work demonstrated that general-purpose LLMs such as
GPT-4 could perform meaningfully on psychiatric knowledge
tasks. A 2024 evaluation published in \textit{The British
Journal of Psychiatry}~\cite{Kim2024bjp} benchmarked five
LLMs --- GPT-4, LLaMA2-70B, Mixtral-45B, Vicuna-13B, and
Gemma-7B --- on 21 clinical case vignettes and 95 DSM-5-TR
multiple-choice questions, finding that GPT-4 outperformed
both psychiatry residents and open-source models on diagnostic
tasks. However, this work relied on zero-shot prompting of
general-purpose models rather than domain-specific fine-tuning,
and all inference was server-side.

Med-PaLM 2~\cite{Singhal2023} demonstrated strong performance
on structured psychiatric assessment tasks, achieving
state-of-the-art results in predicting depression and PTSD
scores from de-identified clinical interview transcripts
and providing DSM-5-aligned diagnostic labels. However,
Med-PaLM 2 is a large proprietary model requiring cloud
infrastructure, with no pathway to on-device deployment.

MentalLLaMA~\cite{Yang2024mentallama} proposed an
interpretable mental health analysis system built on
fine-tuned LLaMA models for social media text, providing
explanatory outputs alongside disorder classifications.
While interpretability is a strength, the work targets
social media data rather than clinical conversations and
does not address privacy-preserving deployment.

MHINDR~\cite{MHINDR2025} proposed a DSM-5 aligned LLM
framework that extracts temporal symptom information from
user-generated text for mental health diagnosis and
intervention recommendation. The framework addresses
temporal dynamics often neglected in prior work, but
operates in a cloud-based inference setting and does not
support clinical conversational input or on-device execution.

The LLM Questionnaire Completion (LMIQ) approach~\cite{LMIQ2024}
demonstrated the potential of LLMs to complete standardized
psychiatric assessment instruments such as the PHQ and PCL-C
from clinical interview transcripts, achieving strong correlation
with clinician-assigned scores. This work establishes the
feasibility of LLM-based structured psychiatric assessment but
does not address deployment constraints or privacy.

Multi-agent LLM frameworks for psychiatric care have also
emerged. A trustworthy AI psychotherapy framework~\cite{TrustworthyAI2025}
proposed a multi-agent LLM workflow for counseling and
explainable mental disorder diagnosis, combining specialized
agent roles for symptom extraction, diagnostic reasoning, and
therapeutic response generation. While the multi-agent design
improves explainability and clinical structure, it requires
cloud-hosted LLM infrastructure.

\subsection{Privacy-Preserving AI in Mental Healthcare}

The tension between the data requirements of AI-based mental
health tools and the privacy expectations of patients and
regulators has motivated a growing body of work on
privacy-preserving approaches.

Federated learning has emerged as the dominant paradigm for
privacy-preserving mental health AI. FedMentalCare~\cite{FedMentalCare2025}
proposed a federated learning framework combining LoRA-based
fine-tuning of LLMs across distributed client nodes, enabling
model training without centralizing patient data and achieving
HIPAA and GDPR compliance. FedMentor~\cite{FedMentor2025}
extended this approach with domain-aware differential privacy
budgets and heterogeneous client support, evaluated on small
mobile-friendly backbones including MobileLLM and SmolLM2.
Federated learning for depression detection from multilingual
social media posts was demonstrated by~\cite{FedDepression2024},
showing that federated models can match centralized baselines
while preserving data locality. A broader review of
privacy-preserving federated learning in healthcare
contexts~\cite{PatiFederated2024} identified residual privacy
risks in gradient exchange and the need for complementary
privacy mechanisms.

However, federated learning preserves privacy during
\textit{training} --- not during \textit{inference}. In
federated systems, the trained model is still typically
deployed on a server that processes patient queries at
inference time, meaning sensitive patient data is still
transmitted for each clinical interaction. The proposed
platform addresses a fundamentally different threat model:
zero-egress inference, in which no patient data is transmitted
at any stage, including at runtime. This distinction makes the
proposed approach particularly suitable for operationally
sensitive environments where data transmission is
architecturally prohibited rather than merely minimized.

\subsection{Mobile and Edge AI for Clinical Decision Support}

The feasibility of deploying capable LLMs on mobile hardware
has been established by several recent works. Laskaridis
\textit{et al.}~\cite{laskaridis2024} systematically evaluated
the inference performance of sub-4B quantized LLMs on consumer
smartphones, finding that models in this range could achieve
5--15 tokens per second on modern devices, sufficient for
interactive clinical use. Xu \textit{et al.}~\cite{xu2024survey}
surveyed resource-efficient LLM architectures for edge
deployment, identifying QLoRA and GGUF quantization as the
most practical approaches for mobile inference. A 2025
measurement study~\cite{MobileEfficiency2025} benchmarked
Gemma-2B and Llama-3.2 on Android smartphones using
MediaPipe and ExecuTorch respectively, providing empirical
latency and memory profiles directly relevant to the proposed
platform's deployment targets.

DRHouse~\cite{DRHouse2024} proposed an LLM-empowered
diagnostic reasoning system that integrates sensor data and
expert knowledge for mobile health applications, demonstrating
the feasibility of combining on-device sensing with LLM
reasoning for clinical inference. However, this work addresses
general diagnostic reasoning rather than psychiatric assessment
specifically, and does not implement a zero-egress privacy model.

A systematic review of LLMs in mobile mental health
applications~\cite{JMIRReview2025} found that while the
majority of LLM-based mental health systems are web or API-based,
local mobile deployment offers advantages in offline access,
device sensor integration, and privacy-aware personalization
--- but noted that as of early 2025, only a small minority of
systems had implemented standalone mobile app deployments.

\subsection{mHealth Platforms for Psychiatric and Clinical Care}

A wide range of mHealth platforms have been developed for
mental health support and clinical decision assistance,
though most rely on cloud-based AI infrastructure.
AI-powered mental health chatbots such as Woebot and Wysa
deliver evidence-based cognitive behavioral therapy
interventions via conversational interfaces, but route all
interactions through cloud servers, limiting their
applicability in privacy-sensitive contexts~\cite{Guo2020}.
A 2025 systematic review of LLM-based systems in mental health
care~\cite{JMIRReview2025} found that accuracy rates for
AI-based depression and anxiety screening reached 96--98\%
for specific metrics, but that data privacy, algorithmic bias,
and the lack of clinical validation remained the primary
barriers to deployment in regulated healthcare settings.

The evolving landscape of digital mental health has been
characterized by a transition from telehealth and
app-based interventions toward generative AI, including
LLM-based conversational agents~\cite{DigitalMH2025}. This
review highlighted that while LLMs introduce unique potential
for psychiatric care, including scalable conversational
assessment and personalizable clinical support, their
deployment raises unresolved questions about regulatory
oversight, clinical validation, and data governance that
app-based solutions have not yet fully addressed.

\subsection{Positioning of the Proposed Work}

Table~\ref{tab:related} summarizes the key dimensions on which
the proposed platform differs from representative prior work.
The proposed platform is, to the best of our knowledge, the
first to combine: (1)~a fine-tuned LLM ensemble specifically
trained on psychiatrist--patient conversational data for
DSM-5-aligned diagnosis; (2)~fully on-device, zero-egress
inference with no data transmitted externally at any stage;
(3)~deployment as a cross-platform mobile clinical decision
support application; and (4)~support for both clinician-facing
differential diagnosis and patient-facing self-screening within
a single privacy-preserving framework.

\begin{table*}[!htb]
\centering
\caption{Comparison of the proposed platform with representative related work.
\cmark~=~supported; \xmark~=~not supported; $\sim$~=~partially supported.}
\begin{adjustbox}{width=1\textwidth}
\label{tab:related}
\begin{tabular}{lcccccc}
\toprule
\thead{Work} &
\thead{Psychiatric\\domain} &
\thead{Fine-tuned} &
\thead{On-device\\inference} &
\thead{Zero-egress} &
\thead{Mobile app} &
\thead{DSM-5\\aligned} \\
\midrule
Med-PaLM 2~\cite{Singhal2023}           & \cmark & \xmark & \xmark & \xmark & \xmark & $\sim$  \\
Kim et al.~\cite{Kim2024bjp}            & \cmark & \xmark & \xmark & \xmark & \xmark & \cmark  \\
MentalLLaMA~\cite{Yang2024mentallama}   & \cmark & \cmark & \xmark & \xmark & \xmark & \xmark  \\
LMIQ~\cite{LMIQ2024}                    & \cmark & \xmark & \xmark & \xmark & \xmark & \cmark  \\
MHINDR~\cite{MHINDR2025}                & \cmark & \cmark & \xmark & \xmark & \xmark & \cmark  \\
FedMentalCare~\cite{FedMentalCare2025}  & \cmark & \cmark & \xmark & \xmark & \xmark & \xmark  \\
FedMentor~\cite{FedMentor2025}          & \cmark & \cmark & $\sim$ & \xmark & \xmark & \xmark  \\
DRHouse~\cite{DRHouse2024}              & \xmark & \xmark & $\sim$ & \xmark & $\sim$ & \xmark  \\
Trustworthy AI~\cite{TrustworthyAI2025} & \cmark & \xmark & \xmark & \xmark & \xmark & \cmark  \\
\midrule
\textbf{Proposed platform}              & \cmark & \cmark & \cmark & \cmark & \cmark & \cmark  \\
\bottomrule
\end{tabular}
\end{adjustbox}
\end{table*}


\section{Conclusion and Future Work}
\label{sec:conclusion}

This paper presented a zero-egress, on-device AI platform for
privacy-preserving psychiatric decision support deployed as a
cross-platform mobile clinical application. The platform
integrates a fine-tuned ensemble of three lightweight LLMs ---
Gemma, Phi-3.5-mini, and Qwen2 --- compressed via QLoRA 4-bit
quantization and coordinated by an on-device consensus
reasoning layer, producing DSM-5-aligned differential diagnoses
for five psychiatric conditions from natural language clinical
conversations without transmitting any patient data externally.
Clinical functionality is delivered through four structured task
flows --- SOAP note generation, ICD-10 coding, clinical research
assistance, and document analysis --- with a dedicated
patient-facing self-screening mode and embedded clinical
safeguards aligned with responsible AI principles.

The key contribution of this work is architectural: the
zero-egress guarantee is enforced at the system level rather
than managed through policy, making the platform uniquely
suitable for operationally sensitive environments such as
military field settings, correctional facilities, and remote
clinics where data transmission is prohibited or constrained.
Preliminary observations confirm on-device inference within a
clinically tolerable latency range, and the structured
comparison with related work establishes that this is, to the
best of our knowledge, the first fully on-device, zero-egress
psychiatric AI platform built on a fine-tuned LLM ensemble and
deployed as a production mobile clinical application.

Future work will focus on three directions: expanding the
\texttt{mental-reasoning} dataset to 5,000--10,000 multilingual
records across additional diagnostic categories and conducting
a prospective clinical validation study with licensed
psychiatrists; running a controlled device-class benchmark study
to establish publication-quality latency, memory, and battery
performance metrics; and pursuing federated fine-tuning for
privacy-preserving model updates~\cite{FedMentalCare2025,
Rieke2020}. A proof-of-concept
clinical deployment is planned in collaboration with McDonald
Army Health Center, Newport News, VA, USA, which will serve as
the primary evaluation site for assessing real-world diagnostic
performance, clinician usability, and platform suitability for
military mental health care in operationally sensitive
environments. Additional directions include multimodal input
via voice and wearable biosensors, longitudinal symptom
tracking, and systematic bias audits across demographic
subgroups~\cite{Chen2021bias}.



\bibliographystyle{elsarticle-num}
\bibliography{reference}

\end{document}